\documentclass[11pt]{article}

\usepackage[final]{acl}

\usepackage{times}
\usepackage{latexsym}
\usepackage[T1]{fontenc}
\usepackage[utf8]{inputenc}
\usepackage{microtype}
\usepackage{inconsolata}
\usepackage{graphicx}
\usepackage{amsmath}
\usepackage{amssymb}
\usepackage{booktabs}

\usepackage{listings}
\title{Controlling Reading Ease with Gaze-Guided Text Generation}

\author{
  Andreas Säuberli\textsuperscript{1,2}\hspace{0.8cm}Darja Jepifanova\textsuperscript{1}\hspace{0.8cm}Diego Frassinelli\textsuperscript{1}\hspace{0.8cm}Barbara Plank\textsuperscript{1,2} \\
  \textsuperscript{1}MaiNLP, Center for Information and Language Processing, LMU Munich, Germany \\
  \textsuperscript{2}Munich Center for Machine Learning (MCML), Munich, Germany \\
  \texttt{andreas.saeuberli@lmu.de}
}

\begin{document}

\maketitle

\begin{abstract}
  The way our eyes move while reading can tell us about the cognitive effort required to process the text. In the present study, we use this fact to generate texts with controllable reading ease. Our method employs a model that predicts human gaze patterns to steer language model outputs towards eliciting certain reading behaviors. We evaluate the approach in an eye-tracking experiment with native and non-native speakers of English. The results demonstrate that the method is effective at making the generated texts easier or harder to read, measured both in terms of reading times and perceived difficulty of the texts. A statistical analysis reveals that the changes in reading behavior are mostly due to features that affect lexical processing. Possible applications of our approach include text simplification for information accessibility and generation of personalized educational material for language learning.
\end{abstract}


\section{Introduction}
\label{sec:introduction}

Controlling the readability of texts is essential for many practical applications. In second language acquisition, teachers need to provide students with material that is simple enough to be understood, but difficult enough to achieve a learning effect \citep{Krashen1985,Nation1999}. Making texts easier to read can also facilitate information access, for example, for people with communication impairments \citep{Maass2020,Karreman2007,Rello2013,Alonzo2020,Saeuberli2024a}.

Recently, natural language processing methods and large language models (LLMs) in particular have been applied to adapt the readability of texts automatically, mostly using prompting techniques \citep{Feng2023,Kew2023,Fang2025}. While these approaches have shown some success, the generated texts are often misaligned with the desired readability levels \citep{Imperial2023,Uchida2025}, and there is a lack of control and transparency in the generation process \citep{Sun2023,Zhao2024}. To counteract some of these limitations, we propose eye-tracking data as a more direct signal for controlling the reading ease of LLM outputs.

The way a person's eyes move while reading a text contains rich information about both the text and the reader \citep{Kuperman2018}. For example, longer reading times on a word reflect greater effort in processing that word \citep{Just1980,Rayner1998}. Consequently, eye tracking has been used as a measure to assess the readability of texts \citep{Rello2013,Singh2016,Vajjala2016,GrutekeKlein2025a}.

\begin{figure}
  \centering
  \includegraphics[width=\columnwidth]{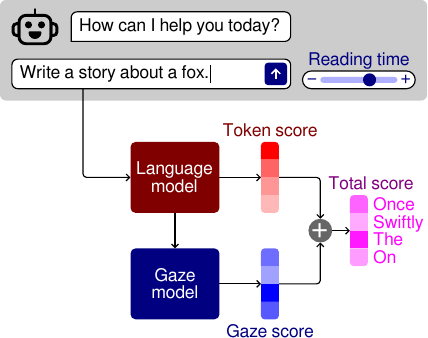}
  \caption{\textbf{Example application of gaze-guided text generation in a chatbot, and overview of the method.} Our approach allows steering language model outputs to elicit certain reading behaviors -- for example, increased or decreased reading time. It integrates a gaze model predicting eye-tracking measures into the decoding process to modify the token probabilities generated by the language model.}
  \label{fig:overview}
\end{figure}

In this paper, we exploit the relationship between eye movements and reading ease from a different perspective. Instead of using gaze data to \emph{measure} the readability of existing texts, we use it to \emph{generate} texts with controllable reading ease. Specifically, we propose a method for guiding language models to produce outputs that will elicit certain reading behaviors, allowing us to steer the generated texts to be more or less easily readable. We achieve this by integrating a \emph{gaze model} into the decoding process (see Figure \ref{fig:overview}). The gaze model predicts eye-tracking measures for candidate tokens generated by the language model. These predictions are then used to re-rank the token candidates. Our human evaluation study demonstrates the success of this method both in terms of the effect on readers' eye movements and their subjective perception of overall difficulty. The contributions of this paper are:

\begin{enumerate}
  \item We introduce a method for using eye-tracking data to guide text generation with language models.
  \item We present a new publicly available eye-tracking-while-reading dataset involving texts generated using this method.
  \item We analyze generated texts, eye-tracking and human rating data to demonstrate the effectiveness of the method.
\end{enumerate}


\section{Related work}

Some controlled text generation approaches rely on fine-tuning a language model to condition the output on a class or attribute given at inference time \citep[e.g.,][]{Keskar2019}. This has several disadvantages: (1) fine-tuning large models is resource-intensive, (2) the method is inflexible because adapting the control mechanism (e.g., adding classes/attributes) requires re-training, and (3) this is often not possible with closed LLMs without access to model parameters. Therefore, several approaches have been introduced to enable \emph{modular} control at inference-time. These approaches (including ours) involve training additional, usually smaller models to steer an off-the-shelf, unmodified language model \citep{Dathathri2020,Krause2021,Liu2021,Arora2022,Li2023b,Liu2024}. Among these, the most similar to our approach is \textsc{Fudge} \citep{Yang2021}, which involves training a binary classifier to predict the desired attribute (e.g., topic or formality) for completion candidates and combining the predictions with the probabilities from the language model. \citet{Kew2022} successfully applied this approach to text simplification using a classifier trained to discriminate between simple and complex sentences. Our approach differs in that we use a regression model to predict continuous variables such as reading times.

While we are the first -- to the best of our knowledge -- to adapt inference-time controlled text generation to eye-tracking data, recent research has proposed alternative methods for integrating eye-movement information into text generation models, for example, by fine-tuning language models directly on gaze data to improve human alignment \citep{LopezCardona2025}. Specifically, \citet{Kiegeland2024a} fine-tuned language models to optimize their alignment to human psychometric data. \citet{Kiegeland2024} used eye-movements as a source of feedback in direct preference optimization for controlled sentiment generation. Finally, \citet{LopezCardona2025a} used synthetic eye-tracking data to model human preferences as a reward for reinforcement learning.

Previous work on the prediction of eye movements while reading include models from computational psycholinguistics such as E-Z Reader \citep{Reichle2003} and SWIFT \citep{Engbert2005}. In natural language processing, various neural network approaches have been applied to predict scanpaths or reading measures, including fine-tuned transformer-based language models \citep{Hollenstein2021c}, recurrent neural networks \citep{Deng2023a}, and diffusion models \citep{Bolliger2023,Bolliger2025a}.


\section{Gaze-guided text generation}
\label{sec:ggtg}

The high-level idea of our method is to manipulate the probability distribution generated by a language model to favor tokens that are likely to elicit a specific reading behavior (e.g., longer/shorter fixations, or more/fewer regressions). It comprises an ensemble of two models:

\begin{enumerate}
  \item a \textbf{language model} that predicts probabilities for the next token, and
  \item a \textbf{gaze model} that predicts an aggregated eye-tracking measure (e.g., reading time or number of regressions) for the sequence generated so far.
\end{enumerate}

Crucially, both models are unidirectional, i.e., predictions are made based on preceding context only, to allow for autoregressive text generation.

\begin{figure*}
  \centering
  \includegraphics[width=0.95\linewidth]{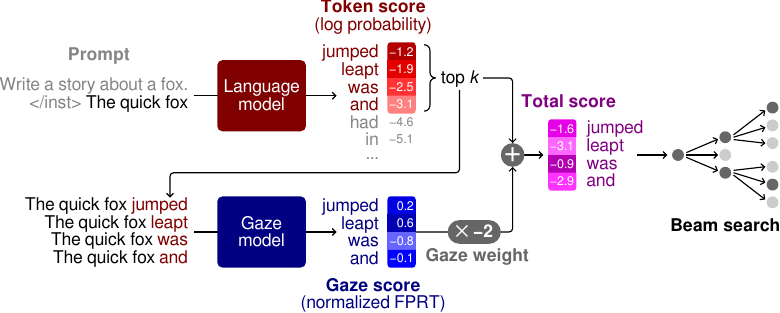}
  \caption{Gaze-guided text generation approach. \textbf{(1)} An off-the-shelf language model predicts a probability distribution (\emph{token scores}) for the next token. \textbf{(2)} A fine-tuned gaze model predicts an eye-tracking measure (\emph{gaze score}; in this case: FPRT, first-pass reading time) for each of the top $k$ candidates. \textbf{(3)} The gaze scores are multiplied with a user-defined \emph{gaze weight} (in this case: $-2$) and added to the token scores. \textbf{(4)} The resulting \emph{total scores} are used for decoding with beam search.}
  \label{fig:method}
\end{figure*}

\pagebreak

The text generation procedure is visualized in Figure \ref{fig:method}. First, we prompt the language model with any user-defined prompt to predict the probability distribution for the next token. We then calculate the log probability for each candidate sequence by summing up the log probabilities for each generated token -- these are the \textbf{token scores}. We find the top $k$ candidates sequences and use the gaze model to predict the eye-tracking measure for each sequence -- the \textbf{gaze scores}. The final \textbf{total scores} are obtained by multiplying the gaze scores with a user-defined \textbf{gaze weight} and adding them to the corresponding token scores. We use the total scores to select the next token through beam search (or any other decoding strategy).

To exemplify this, consider a gaze model that predicts reading time as our gaze score. In this case, a positive gaze weight means that tokens that yield long reading times will be favored over tokens that yield short reading times. On the contrary, a negative gaze weight (as in Figure \ref{fig:method}) will favor tokens with shorter reading times, thereby potentially enhancing reading ease. A gaze weight of 0 corresponds to disabling the gaze model and only using the token scores from the language model.

The modular nature of this setup means that both the language model and the gaze model can be independently replaced without retraining the other model. However, since the two models may require different tokenizers, there is a chance that partial words generated by the language model do not match the gaze model's tokenization, leading to unreliable gaze scores for those subwords. In our case, we alleviated this problem by applying beam search (with beam size $k$). This prevents the model from relying too much on gaze score predictions of subwords that might not be part of the gaze model's vocabulary, and allows it to `revise' previously generated tokens as the gaze model's tokenization of the final word stabilizes.\footnote{A similar issue arises in token-level language model ensembling. See \citet{Chen2025} for a discussion of more advanced solutions in this context.}

For our experiments, we chose word-level \textbf{first-pass reading time} (FPRT; sometimes referred to as \emph{gaze duration}) as the eye-tracking measure, which is the sum of fixation durations on a word before the gaze moves away from the word for the first time (see Section \ref{sec:experiment-gaze-model}). Since the gaze model is required to predict a single value for each candidate sequence, we use the sum of predicted word-level FPRTs across the entire candidate sequence as our gaze score.


\section{Experimental setup}

The goal of our study is to measure the effect of gaze-guided text generation, reading behavior, and subjective perception by human readers. To this end, we generated texts with different gaze weights and presented these texts to L1 and L2 readers, tracking their eye movements while reading and collecting ratings from them.

\subsection{Text generation}

We opted to generate short fictional stories to minimize individual differences in participants' topic familiarity and to increase motivation. All the texts were in English due to the availability of eye-tracking data for this language, which was required to train the gaze model.

\begin{table*}
  \centering
  \resizebox{\textwidth}{!}{
    \begin{tabular}{p{1cm}p{15cm}}
      \toprule
      \raggedleft Gaze weight & Output \\
      \midrule
      \raggedleft $-5$ & \small Emily had always been the girl who came in second. It was as if she had a magnet on her back that drew all the best things in life to the people in front of her. In school, she was always the one who came in second in the class, the one who got the second prize in the science fair, and the one who lost to her best friend in every game of tag. [...] \\
      \raggedleft $-2$ & \small For as long as she could remember, Emma had come in second place. It wasn't just in school or on the sports field -- it was in every aspect of her life. Whether it was a contest, a competition, or even a relationship, she always seemed to fall just short of the top spot. [...] \\
      \raggedleft $0$ & \small Emily had always been the bridesmaid, never the bride. In school, she was the runner-up in every science fair, the second-place winner in every local art competition, and the silver medalist in every regional sports tournament. [...] \\
      \raggedleft $2$ & \small Emma had always been the perpetual bridesmaid, the perpetual runner-up, the perpetual almost-but-not-quite. It wasn't just limited to contests, either. Relationships, friendships, even everyday accomplishments -- she consistently found herself finishing second. [...] \\
      \raggedleft $5$ & \small Emma's perpetual existence revolved around mediocrity. Every endeavor, every competition, every social interaction, she consistently ranked second-best-in-the-back-of-the-pack-to-last-but-not-quite-second-place-but-closer-than-the-last-place-[...] \\
      \bottomrule
    \end{tabular}
  }
  \caption{Sample outputs generated with different gaze weights. Note the differences in word length, lexical complexity, and the repetitive failure mode at \emph{gaze weight 5}.}
  \label{tab:outputs}
\end{table*}

\subsubsection{Language model}

We used the off-the-shelf instruction-tuned Llama 3.2 model with 3 billion parameters \citep{Grattafiori2024} due to its strong performance relative to its size and speed. The user messages for the language model included a story title and prompt, and the instruction to generate a short story of around 500 words. To come up with the story prompts, we used GPT-4o (OpenAI, April 2025)\footnote{\url{https://chatgpt.com/}} and manually selected six prompts to represent a variety of topics. All story and instruction prompts can be found in Appendix \ref{sec:appendix-prompts}.

\subsubsection{Gaze model}
\label{sec:experiment-gaze-model}

As mentioned in Section \ref{sec:ggtg}, we used word-level FPRT as the eye-tracking measure for the gaze model. The reason for this choice is that (1) FPRT is commonly used in psycholinguistic studies to measure the effort involved in early cognitive processes \citep[pp.\ 221--222]{Godfroid2019}, and (2) it can be plausibly predicted based on preceding context only, since it only includes fixations made during the first pass over the word.\footnote{In preliminary experiments, we also explored other eye-tracking measures such as go-past time, which includes regression path durations, but found that the gaze model performed substantially worse for these measures (see Appendix \ref{sec:appendix-gaze-model}).}

We fine-tuned a GPT-2 model with 124 million parameters \citep{Radford2019} to predict $z$-score-normalized FPRT for each word. The training data is taken from the EMTeC dataset \citep{Bolliger2025a}, which contains eye movements of 107 L1 English speakers recorded while reading texts generated by language models with various decoding strategies. This makes it an ideal dataset for our use case.

Mapping word-level reading times to subword-level tokens is a problem inherent to the processing of eye-tracking data with modern language models \citep[p.\ 43]{Beinborn2024}. To solve this issue in our training data, we distributed each word's FPRT uniformly across all its subwords. At inference time, the sum of predicted FPRTs over all tokens represents our gaze score.

The resulting model achieved 35.5\% explained variance ($R^2$) on a held-out set of 20\% of texts, outperforming a linear regression baseline with word length and frequency as features ($R^2$ = 31.4\%). For more details on the model and the training process, refer to Appendix \ref{sec:appendix-gaze-model}.

\subsubsection{Decoding and gaze weights}

Based on preliminary exploration, we used the top $k = 8$ candidates for gaze prediction and applied beam search with a beam size of 8 for decoding.

For each story prompt, we generated texts for several gaze weights between $-5$ (decrease FPRT) and $+5$ (increase FPRT). For our chosen language model, gaze weights beyond $\pm3$ tended to cause repetitive failure modes in the output (see Section \ref{sec:results-text} and Figure \ref{fig:text-stats}). Therefore, we selected the gaze weights $-2$, $0$, and $+2$ for measuring the effect of gaze guidance in the eye-tracking study. Table \ref{tab:outputs} contains sample outputs for different gaze weights.

\subsection{Eye-tracking study}

In a new eye-tracking study, we recorded participants' eye movements while reading the texts generated with gaze weights $-2$, $0$, and $+2$ (a total of 18 texts). We recruited both L1 and L2 speakers of English, since the gaze model was trained on L1 data, while one of the main applications of the approach is L2 language learning.

\subsubsection{Participants}

Twelve participants in the L1 group and twelve participants in the L2 group were recruited primarily through university mailing lists and courses. Informed consent was obtained before conducting the study. Participants received 10 euros or participated as part of a university course.

Participants' ages ranged between 19 and 53 years (mean: 27.2$\pm$7.0). Participants were categorized as L1 when they self-reported that English was the first language they learned, or that they grew up multilingually with English and had native-level proficiency. In the L1 group, reported language varieties were American (5), British (5), Indian (1), and Singeporean English (1). In the L2 group, native languages included German (8), Chinese (2), Russian (1) and Turkish (1), while self-reported English proficiency levels ranged from C1 (9) to C2 (3) in the Common European Framework of Reference (CEFR).

\subsubsection{Apparatus}

We recorded participants' eye movements with an \emph{EyeLink 1000 Plus} eye tracker (SR Research, Kanata, Ontario, Canada) with a sampling rate of 1000 Hz. Only the dominant eye was tracked, which was determined in a test at the start of the session. We used a chin and forehead rest to minimize head movements and to maintain a constant eye-screen distance of 66 cm. The stimuli were presented on a 24.5-inch monitor with a resolution of 1920$\times$1080 pixels and a refresh rate of 240 Hz. All content was presented within 28$\times$19 cm or \textasciitilde25$\times$17 degrees of visual angle. We performed 9-point calibration at the beginning and, if necessary, between texts. Texts were typeset in \emph{Noto Sans}\footnote{\url{https://notofonts.github.io}} in black on a light gray background, with an x-height of 4 mm or \textasciitilde0.35 degrees of visual angle and double line spacing. We used \emph{PsychoPy} \citep{Peirce2019} to implement the stimulus presentation software.

\subsubsection{Procedure}

Data was collected in a single session of 30--45 minutes per participant. After reading instructions on the screen and a practice trial, participants read six texts (two for each gaze weight). Each participant saw one text for each of the six story prompts. Depending on the length, the texts were presented on 4 or 5 pages. Participants could advance to the next page by pressing a key on a keyboard, but could not return to previous pages. Each text was followed by a multiple-choice comprehension question for attention checking and three 4-point Likert-style ratings on the difficulty, naturalness/fluency, and interestingness of the text. The texts were assigned such that each participant saw exactly one story for each prompt, and the order was randomized for each participant.

\subsubsection{Data cleaning and processing}

We follow common practice in performing post-hoc corrections of eye-tracking data \citep[pp.\ 257--259]{Godfroid2019}. We manually defined thin-plate spline transformations on the raw data to correct vertical misalignment with the text lines where necessary and possible. Trials or parts of trials with very poor quality were removed.

The participants' answers to the comprehenion questions indicated that the texts were read attentively overall. All participants achieved at least 5 out of 6 correct answers (average: 94.4\% accuracy overall). Therefore, we did not remove any additional trials on this basis.

We then used \emph{pymovements} \citep{Krakowczyk2023} to process the cleaned data. We applied the I-DT algorithm for fixation detection \citep{Salvucci2000} with a dispersion threshold of 1.0 degrees of visual angle and a minimum fixation duration of 100 ms.

In accordance with \citet{Jakobi2024}, we report more detailed data quality metrics in Appendix \ref{sec:appendix-data-quality} and publish both the raw and the cleaned and preprocessed versions.


\section{Results}

We examine the effects of gaze-guided text generation from three perspectives:

\begin{enumerate}
  \item \textbf{Effects on text characteristics:} How does changing the gaze weight affect linguistic features in the generated texts?
  \item \textbf{Effects on reading behavior:} How does changing the gaze weight affect first-pass reading times in L1 and L2 readers?
  \item \textbf{Effects on subjective perception:} How does changing the gaze weight affect perceived difficulty, naturalness, and interestingness in L1 and L2 readers?
\end{enumerate}

\subsection{Effects on text characteristics}
\label{sec:results-text}

\begin{figure}
  \centering
  \includegraphics[width=\columnwidth]{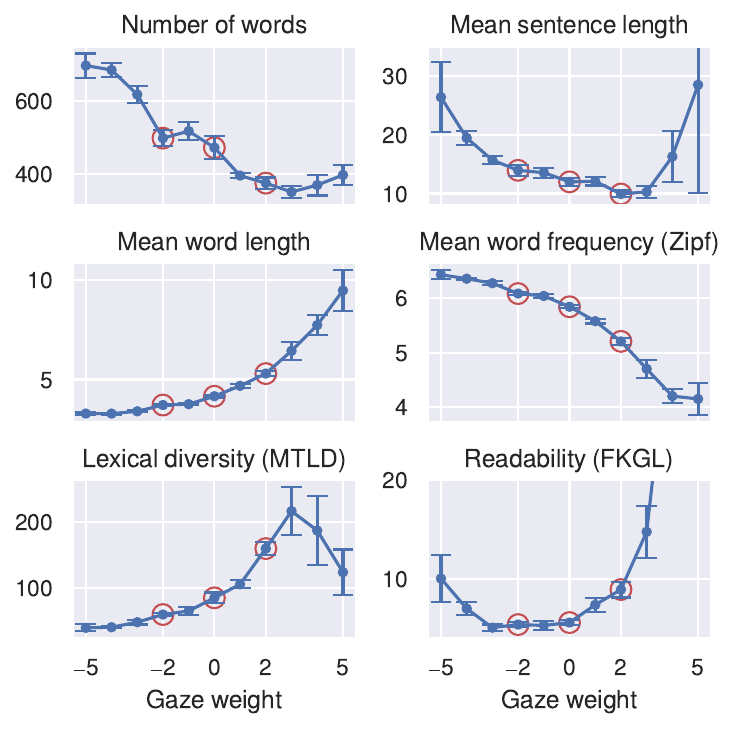}
  \caption{Text characteristics across different gaze weights. Error bars show standard error of the mean from six generated texts. Points circled in red represent the texts shown to participants in the eye-tracking study.}
  \label{fig:text-stats}
\end{figure}

We calculated word-level, sentence-level, and document-level statistics across different gaze weights, including word frequency (Zipf score, based on the Python package \emph{wordfreq}\footnote{\url{https://pypi.org/project/wordfreq/}}, \citealp{Speer2022}), lexical diversity (measure of textual lexical diversity [MTLD], \citealp{McCarthy2005}; Python package \emph{lexical-diversity}\footnote{\url{https://pypi.org/project/lexical-diversity/}}), and readability (Flesch-Kincaid grade level [FKGL], \citealp{Kincaid1975}; Python package \emph{textstat}\footnote{\url{https://pypi.org/project/textstat/}}). Figure \ref{fig:text-stats} shows several trends. Increasing the gaze weight tends to lead to texts with fewer but longer and less frequent words, shorter sentences, more lexical diversity, and a higher reading level. Gaze weights beyond $\pm3$ sometimes produced highly repetitive texts, resulting in extreme values in the case of sentence length, lexical diversity, and readability. This degeneration in output quality is somewhat expected -- with larger gaze weights, the gaze model overpowers the language model's output probabilities, potentially assigning high probability to contextually implausible words. In preliminary experiments, smaller language models were more prone to this behavior.

\subsection{Effects on reading behavior}
\label{sec:results-reading}

\begin{figure}
  \centering
  \includegraphics[width=\columnwidth]{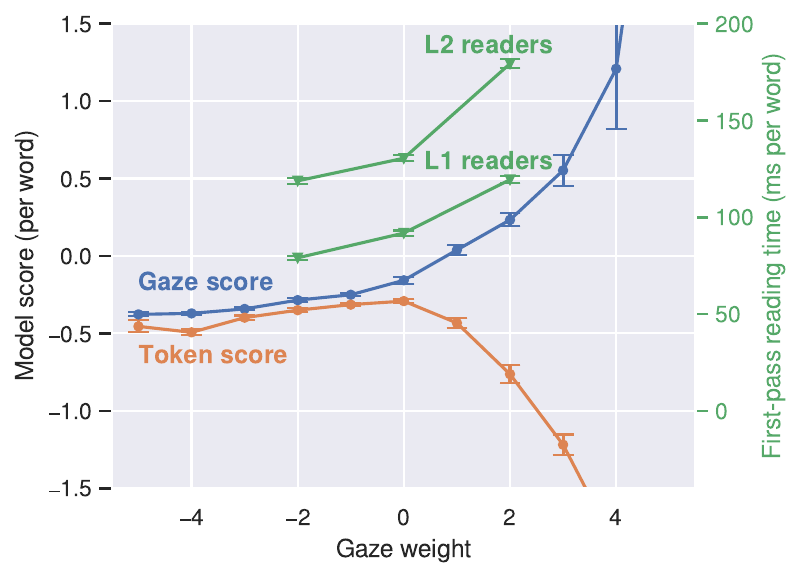}
  \caption{Length-normalized gaze and token scores predicted by the models, compared to first-pass reading times measured in the eye-tracking study. Token scores are token log probabilities predicted by the language model. Gaze scores are normalized first-pass reading times predicted by the gaze model. Error bars show standard error of the mean.}
  \label{fig:score-fprt}
\end{figure}

In Figure \ref{fig:score-fprt}, we compare the gaze and token scores predicted by the models with the observed reading times in the eye-tracking experiment (recall that readers saw texts with gaze weights $\pm2$). As expected, gaze scores increase with positive gaze weights and decrease with negative gaze weights, while the token score is maximal when no gaze model is used. FPRT measurements roughly follow the predicted gaze scores, although the effect is stronger for L2 readers compared to L1 readers. A linear mixed-effects model with fixed effects for positive and negative gaze weights and random intercepts for readers confirms significant effects for the gaze weights in both reader groups (all $p < 0.001$).

\begin{figure}
  \centering
  \includegraphics[width=\columnwidth]{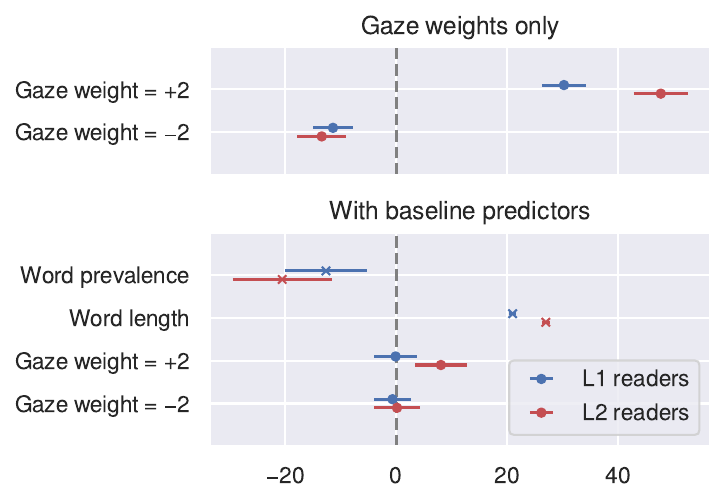}
  \caption{Coefficients and 95\% confidence intervals of linear mixed-effects models fitted to observed first-pass reading times. \textbf{Upper panel:} Model with gaze weight as predictor (compared to the reference with gaze weight 0). \textbf{Lower panel:} Model with word length and prevalence as additional baseline predictors. Both models include random intercepts for readers.}
  \label{fig:model-coef}
\end{figure}

Psycholinguistic studies have established that simple features like word length or prevalence (i.e., the proportion of native speakers who know the word) are highly predictive of processing effort \citep{Barton2014,Brysbaert2016}. To determine how much these features contributed to the differences in reading time, we fitted additional models with word length (number of characters) and prevalence (according to \citealp{Brysbaert2018}\footnote{Words that are not part of \citeauthor{Brysbaert2018}'s (\citeyear{Brysbaert2018}) list were excluded from the analysis. This amounted to an exclusion of 3.6\% of all data points.}) as additional fixed effects. As shown by Figure \ref{fig:model-coef}, we find negative effects for word prevalence (i.e., well-known words are read more quickly) and positive effects for word length (i.e., longer words are read more slowly), as expected. However, adding these baseline predictors also mostly renders the gaze weight effects insignificant, with the exception of positive gaze weight in L2 readers. In other words, in the L1 case, differences in word length and prevalence fully explain the changes in reading time induced by the gaze guidance, whereas in the L2 case, increasing the gaze weight led to increases in reading time \emph{beyond} length and prevalence effects. Overall, this shows that the gaze model mostly captures features related to lexical processing. More details on the mixed-effects models can be found in Appendix \ref{sec:appendix-mixed-effects-models}.

\subsection{Effects on subjective perception}
\label{sec:results-ratings}

\begin{figure}
  \centering
  \includegraphics[width=\columnwidth]{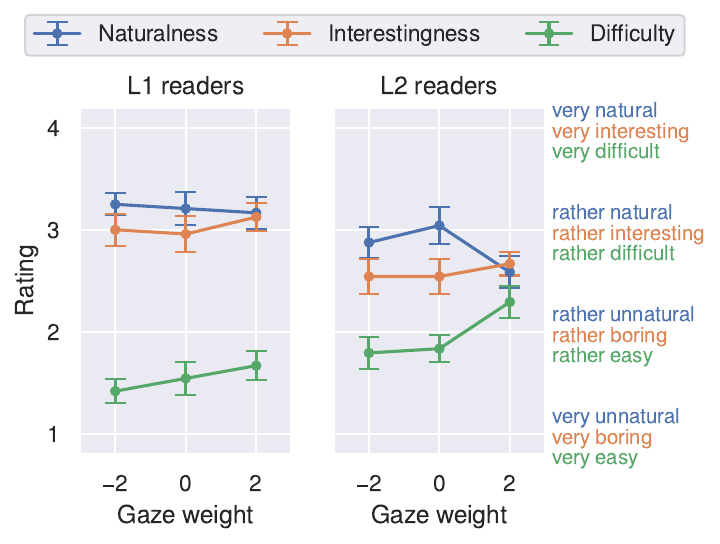}
  \caption{Mean difficulty, naturalness, and interestingness ratings across gaze weights. Error bars show standard error of the mean.}
  \label{fig:ratings}
\end{figure}

Mean difficulty, naturalness, and interestingness ratings are visualized in Figure \ref{fig:ratings}. In both groups, we observe a consistent trend of rising difficulty ratings with increasing gaze weights, and a substantial increase in difficulty for L2 readers. L2 readers rated texts with a positive gaze weight as less natural, likely reflecting the increased perceived difficulty as opposed to poor fluency. Interestingness remained largely unaffected in both groups.

These results show that readers perceived an increase in difficulty with larger gaze weights, and that the gaze guidance did not substantially affect text quality.

\subsection{Gaze model performance}

\begin{figure}
  \centering
  \includegraphics[width=0.9\columnwidth]{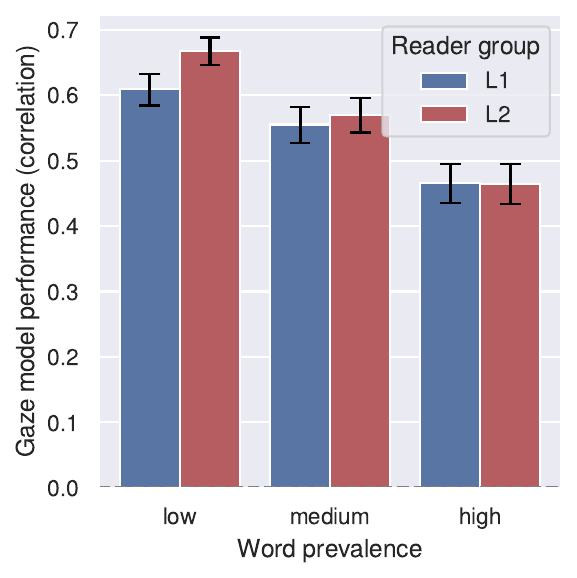}
  \caption{Pearson correlation between word-level gaze model predictions and observed first-pass reading time in the eye-tracking study. Error bars represent 95\% confidence intervals. The low, medium, and high prevalence buckets are equally sized quantiles.}
  \label{fig:gaze-model-eval}
\end{figure}

To analyze the gaze model's performance in more depth, we consider how well its word-level predictions on the six generated texts correlate with the observed reading times in our eye-tracking study. The Pearson correlation coefficient for the L2 group is slightly higher than for the L1 group ($r = 0.585$ vs.\ $r = 0.552$). For comparison, the correlation on the held-out data split during training is $r=0.596$. Figure \ref{fig:gaze-model-eval} shows that the correlation the model also performs better with low-prevalence words.

\section{Discussion}

\subsection{Effectiveness of gaze guidance}

Based on the results above, we conclude that our method is successful in adapting language model outputs to different levels of reading ease without impacting text quality. The approach was less effective for reducing FPRT than for increasing it. This is easily explained by the fact that there is a lower bound, but not an upper bound on processing effort in humans -- in other words, making a text more difficult to read is always possible and easier to do, while simplification is only possible to a limited degree and more challenging to achieve.

An important limitation revealed by our experiments is that the effects introduced by the gaze model mostly reflect low-level lexical information like word length and prevalence. This can be attributed to our choice of eye-tracking measure. FPRT and other measures based on fixation durations are known to be strongly influenced by these local, word-level features \citep{Kliegl2006}, while measures like rereading time are commonly associated with later cognitive processes at the level of syntax \citep[p.\ 225]{Godfroid2019} and may therefore be better suited as a proxy for sentence-level reading ease \citep{Vajjala2016}. However, reliably predicting these measures with a gaze model is more difficult and implausible with a unidirectional language model. Therefore, future research could improve our approach by integrating more complex generative models of eye movements \citep[e.g.][]{Deng2023a,Bolliger2023,Bolliger2025} and using a decoding strategy that supports more backtracking to revise previously generated tokens.

\subsection{Differences between L1 and L2 readers}

Overall, we see similar effects for the two reader groups, apart from expected differences like faster overall reading time and lower perceived difficulty for L1 readers.

The most notable difference is the observation that word length and prevalence effects cannot entirely explain the increase in L2 readers' FPRT on texts with gaze weight 2. One possible reason for this could be that the prevalence values are based on native speakers' vocabulary knowledge, while the L2 readers in our study may have encountered a large number of unknown words in those texts, which could affect reading times beyond the linear effect of prevalence. This would also explain the sharp increase in difficulty ratings and the lower naturalness ratings for texts with gaze weight 2. Another possibility could be higher complexity at the syntactic level, although we could not find direct evidence for this.


\section{Conclusion and future work}

We presented a method for generating texts with an inference-time mechanism to control reading ease. The approach is modular: the language model does not require fine-tuning, and the gaze model can be easily replaced, making it both efficient and flexible. We conducted an eye-tracking experiment to assess the effects on readers' cognitive processing effort and perceived difficulty, and release the collected dataset for further research. The results demonstrated that the approach is successful in steering the language model towards more or less easily readable texts, and that this effect is primarily due to changes in lexical complexity.

Manipulating reading ease is of use in educational scenarios for language learning, and in text simplification to enhance communication accessibility. Both of these applications involve users with diverse needs, and can thus benefit substantially from personalization \citep{Bingel2018,Tetzlaff2020}. Eye movements while reading can capture -- at least to some degree -- differences in how well readers understand a text, how they approach a reading task, as well as information on language knowledge and reading difficulties \citep{Koornneef2016,Ahn2020,Raatikainen2021,Skerath2023,Shubi2024,Shubi2025}. While the gaze model in the present work was trained on a general group of L1 speakers, the training data can be easily replaced to reflect a different user group, provided that sufficient eye-tracking data exists for that group. Thus, through further research, gaze-guided text generation could enable personalized generation of educational or informational material, adapted to the reading behavior of the specific reader (or reader group) at hand. Our work represents a first step in this direction.


\section*{Data availability and reproducibility}

The collected dataset, including raw and processed eye-tracking data, is publicly available for research purposes through this link: \url{https://osf.io/rhgbk}. The dataset is also integrated into the \emph{pymovements} Python library\footnote{\url{https://pymovements.readthedocs.io/}} (\citealp{Krakowczyk2023,Krakowczyk2025}) for easy access.

All datasets and code libraries used in this project are open-source and received due citations. The code and data for reproducing the results and figures in this paper is available from this repository: \url{https://github.com/mainlp/gaze-guided-text-generation/}.


\section*{Limitations}

\paragraph{English only.} We only performed experiments with English texts, limiting conclusions about other languages. The main reason for this is that the dataset we used for training the gaze model only contains English, and that eye-tracking-while-reading data is generally sparse in other languages.

\paragraph{Model selection.} We only tested one combination of language and gaze models. We restricted our experiments to this basic setup due to the resources required for collecting eye-tracking data. Testing more models would require multiple sessions for each participant, or recruiting a larger number of participants.

\paragraph{Downstream applicability.} Our experiments only evaluated the approach with the FPRT reading measure, which mostly captures low-level lexical features. This limits the use for text simplification applications, where simplification of syntactic, semantic and discourse-level aspects is usually required. While we believe that our method can also be applied to reading measures that reflect these aspects, this has not been shown empirically in the present work.

\paragraph{User groups.} Although we list text simplification in the context of accessibility as a potential use case, we did not include primary user groups for this application. We were careful not to make any statements about our method's usefulness for communication accessibility, but this remains an important limitation of our work.


\section*{Ethical considerations}

\paragraph{Eye-tracking data.}

Experiments involving human participants require careful consideration. Our study participants took part on a voluntary basis, either as part of a university course or as paid participants (10 euros for a 30--45 min session). We obtained informed consent from participants and allowed them to withdraw their participation at any point. We provided detailed information to participants beforehand. We only publish fully pseudonymized data. While eye-tracking data is considered biometric and identifying individuals from it is technically possible \citep{Jaeger2020}, doing so would require disproportionate effort.

\paragraph{Usage of generative models.}

We used GitHub Copilot for coding assistance, and ChatGPT to support writing the story prompts and comprehension questions. All generated content was thoroughly checked and tested. We did not use generative models for interpreting the results or writing this paper.


\section*{Acknowledgements}

We would like to thank the three anonymous reviewers, whose constructive comments contributed significantly to the final paper. We are also grateful to members of MaiNLP lab, particularly Silvia Casola and Shijia Zhou, as well as Yang Janet Liu for helpful feedback on early drafts of the manuscript.

This research is in parts supported by the ERC Consolidator Grant DIALECT 101043235.


\bibliography{references}

\clearpage
\appendix

\section{Language model prompts}
\label{sec:appendix-prompts}

We used the user message below to prompt the language model. The system prompt is the default defined in the Hugging Face implementation for \texttt{meta-llama/Llama-3.2-3B}.

\begin{lstlisting}
  Write a short story based on the following title and prompt.
  Title: {title}
  Prompt: {prompt}

  The story should not be longer than 500 words
  Keep in mind that the reader will not see the prompt, only the story itself.
  Do not include the title.
\end{lstlisting}

The specific titles and prompts substituted into this template are listed below. 

\begin{enumerate}
    \item \textbf{The Forgotten Message:} A character finds an old voicemail on their phone from a number they don’t recognize. The message is short, cryptic, and chilling.
    \item \textbf{Breakfast with a Stranger:} While eating at a local diner, a character is mistakenly given someone else's breakfast order—and a note tucked under the plate that simply reads, “It’s time.
    \item \textbf{One Last Wish:} A child claims their elderly pet goldfish granted them a wish before it died. The next day, something unexpected happens.
    \item \textbf{Delayed Flight:} Two strangers, both stuck in an airport during a storm, strike up a conversation that changes the course of their lives.
    \item \textbf{Second Place:} A lifelong second-place finisher in everything—from contests to relationships—finally wins something. But the prize comes with a catch.
    \item \textbf{The Last Light On:} In a neighborhood blackout, only one house still has power — but no one seems to live there.
\end{enumerate}

\section{Gaze model training details}
\label{sec:appendix-gaze-model}

\paragraph{GPT-2 fine-tuning:}
\begin{itemize}\setlength\itemsep{0pt}
  \item Base model checkpoint: \\ \texttt{openai-community/gpt2}
  \item Regression head: 1 linear layer
  \item Loss function: mean squared error
  \item Batch size: 10
  \item Learning rate: $1 \times 10^{-4}$
  \item Early stopping patience: 3 epochs
  \item Optimizer: AdamW
\end{itemize}
    
\paragraph{Linear regression baseline:}
\begin{itemize}\setlength\itemsep{0pt}
  \item Features:
  \begin{itemize}
    \item Word length (in characters) of current word and two previous words
    \item Word frequency (according to \emph{wordfreq}; \citealp{Speer2022}) of current word and two previous words
  \end{itemize}
  \item Implementation: \emph{scikit-learn} \citep{Pedregosa2011}
\end{itemize}

Evaluation metrics on the held-out 20\% test set are listed in Table \ref{tab:gaze-model-evaluation}.

\begin{table}[h]
  \centering
  \begin{tabular}{llll}
    \toprule
    & \multicolumn{3}{c}{First-pass reading time} \\
    Model & MSE \textdownarrow & MAE \textdownarrow & $R^2$ \textuparrow \\
    \midrule
    Transformer & 0.631 & 0.597 & 0.355 \\
    Linear regression & 0.671 & 0.615 & 0.314 \\
    \bottomrule
    \toprule
    & \multicolumn{3}{c}{Go-past time} \\
    Model & MSE \textdownarrow & MAE \textdownarrow & $R^2$ \textuparrow \\
    \midrule
    Transformer & 0.715 & 0.576 & 0.292 \\
    Linear regression & 0.843 & 0.600 & 0.165 \\
    \bottomrule
  \end{tabular}
  \caption{Evaluation metrics for the two gaze models.}
  \label{tab:gaze-model-evaluation}
\end{table}

\section{Eye-tracking data quality}
\label{sec:appendix-data-quality}

\begin{itemize}\setlength\itemsep{0pt}
  \item Average calibration accuracy: 0.50° \\ (min: 0.29°, max: 0.82° across participants)
  \item 7/144 trials removed \\ 12/144 trials partially removed
  \item 2.47\% data loss in raw data (e.g., blinks) \\ (min: 0.47\%, max: 16.30\% across participants)
  \item 4.42\% data loss in cleaned data (including manually removed areas) \\ (min: 0.07\%, max: 38.90\% across participants)
  \item Average comprehension question accuracy: 94.4\% \\ (min: 5/6, max: 6/6 questions across participants)
\end{itemize}

\onecolumn
\section{Linear mixed-effects model details}
\label{sec:appendix-mixed-effects-models}

We used \emph{statsmodels} \citep{Seabold2010} to fit the linear mixed-effects models. Table \ref{tab:model-coef} shows the model specifications and coefficients for both reader groups. In the model formulas, \texttt{fprt} is the first-pass reading time on the area of interest (word), \texttt{gaze\_weight} is a categorical variable with three levels (reference/$0$, positive/$+2$, and negative/$-2$), \texttt{word\_length} is the number of characters in the area of interest (including punctuation), \texttt{word\_prevalence} is the word's prevalence value according to \citet{Brysbaert2018}, and \texttt{reader} is the reader's ID.

\begin{table*}[h]
  \centering
  \resizebox{\textwidth}{!}{
    \begin{tabular}{lllllll}
      \toprule
      Model
      & Intercept
      & \texttt{gaze\_weight[+2]}
      & \texttt{gaze\_weight[-2]}
      & \texttt{word\_length}
      & \texttt{word\_prevalence}
      & $R^2$
      \\
      \midrule
      L1 readers: \\
      \raggedright\texttt{fprt \textasciitilde\ gaze\_weight + (1|reader)}
      & $90.1\ _{78.9}^{101.3}$
      & $30.3\ _{26.2}^{34.3}$
      & $-11.4\ _{-15.0}^{-7.8}$
      & ---
      & ---
      & 0.036
      \\
      \addlinespace
      \raggedright\texttt{fprt \textasciitilde\ word\_length + word\_prevalence + (1|reader)}
      & $26.8\ _{6.8}^{46.7}$
      & ---
      & ---
      & $21.0\ _{20.4}^{21.6}$
      & $-12.7\ _{-20.0}^{-5.4}$
      & 0.179
      \\
      \addlinespace
      \raggedright\texttt{fprt \textasciitilde\ gaze\_weight + word\_length + word\_prevalence + (1|reader)}
      & $27.1\ _{6.9}^{47.2}$
      & $-0.1\ _{-3.9}^{3.7}$
      & $-0.7\ _{-4.0}^{2.6}$
      & $21.0\ _{20.4}^{21.6}$
      & $-12.7\ _{-20.0}^{-5.3}$
      & 0.179
      \\
      \midrule
      L2 readers: \\
      \raggedright\texttt{fprt \textasciitilde\ gaze\_weight + (1|reader)}
      & $129.9\ _{112.9}^{146.9}$
      & $47.7\ _{42.8}^{52.6}$
      & $-13.4\ _{-17.9}^{-9.0}$
      & ---
      & ---
      & 0.054
      \\
      \addlinespace
      \raggedright\texttt{fprt \textasciitilde\ word\_length + word\_prevalence + (1|reader)}
      & $63.3\ _{37.7}^{88.9}$
      & ---
      & ---
      & $27.4\ _{26.7}^{28.1}$
      & $-22.3\ _{-31.2}^{-13.4}$
      & 0.206
      \\
      \addlinespace
      \raggedright\texttt{fprt \textasciitilde\ gaze\_weight + word\_length + word\_prevalence + (1|reader)}
      & $59.1\ _{33.3}^{85.0}$
      & $8.1\ _{3.4}^{12.7}$
      & $0.1\ _{-3.9}^{4.2}$
      & $27.0\ _{26.2}^{27.7}$
      & $-20.6\ _{-29.5}^{-11.6}$
      & 0.206
      \\
      \bottomrule
    \end{tabular}
  }
  \caption{Linear mixed-effects model formulas, effects with 95\% confidence intervals, and explained variance (conditional $R^2$).}
  \label{tab:model-coef}
\end{table*}

\end{document}